\documentclass[11pt]{article}

\usepackage[final]{acl}
\usepackage{times}
\usepackage{latexsym}
\usepackage{subcaption}
\usepackage{multirow}
\usepackage{array}
\usepackage{booktabs}
\usepackage{mathptmx}
\usepackage{amssymb}
\usepackage{xfrac}
\usepackage{amsmath}
\usepackage{amsfonts}
\usepackage{amsthm}
\usepackage{makecell}
\usepackage{cleveref}
\usepackage{enumitem}
\usepackage{fancyvrb}
\usepackage{textcomp}
\usepackage[T1]{fontenc}
\usepackage[utf8]{inputenc}
\usepackage{microtype}
\usepackage{inconsolata}

\usepackage{graphicx}
\usepackage{xcolor}
\usepackage{colortbl}
\usepackage{fontawesome}

\newcommand\sect[1]{\S\ref{#1}}
\definecolor{melon}{HTML}{F89E7B}
\definecolor{lavender}{HTML}{967BB6}
\definecolor{teald}{HTML}{008080}

\definecolor{steelblue}{RGB}{60, 130, 190}   
\definecolor{burntorange}{RGB}{245, 130, 45} 
\newcommand{\rparagraph}[1]{\vspace{1.2mm}\noindent\textbf{#1.}}
\usepackage[separate-uncertainty = true,
  multi-part-units = repeat]{siunitx}[=v2]
\title{Sequence Repetition Enhances Token Embeddings\\and Improves Sequence Labeling with Decoder-only Language Models}

\author{Matija Luka Kuki\'{c}$^*$ \quad Marko \v{C}uljak$^{*\dagger}$ \quad David Duki\'{c}$^*$ \quad Martin Tutek \quad Jan \v{S}najder \\
  TakeLab @ Faculty of Electrical Engineering and Computing \\
  University of Zagreb \\
  \texttt{mculjak@fer.hr}}

\begin{document}
\maketitle
\def\thefootnote{*}\footnotetext{Equal contribution, order determined by ascending age}\def\thefootnote{\arabic{footnote}}
\def\thefootnote{$\dagger$}\footnotetext{Corresponding author}
\def\thefootnote{\arabic{footnote}}
\setcounter{footnote}{0}

\def\sectionautorefname{Appendix}
\def\subsectionautorefname{Appendix}
\def\subsubsectionautorefname{Appendix}
\begin{abstract}
Modern language models (LMs) are trained in an autoregressive manner, conditioned only on the prefix. In contrast, sequence labeling (SL) tasks assign labels to each individual input token, naturally benefiting from bidirectional context. This discrepancy has historically led SL to rely on inherently bidirectional encoder-only models. However, the rapid development of decoder-only models has raised the question of whether they can be adapted to SL. While causal mask removal has emerged as a viable technique for adapting decoder-only models to leverage the full context for SL, it requires considerable changes to the base model functionality. In this work, we explore sequence repetition (SR) as a less invasive alternative for enabling bidirectionality in decoder-only models.
Through fine-tuning experiments, we show that SR inherently makes decoders bidirectional, improving the quality of token-level embeddings and surpassing encoders and unmasked decoders.
Contrary to earlier claims, we find that increasing the number of repetitions does not degrade SL performance. Finally, we demonstrate that embeddings from intermediate layers are highly effective for SR, comparable to those from final layers, while being significantly more efficient to compute. Our findings underscore that SR alleviates the structural limitations of decoders, enabling more efficient and adaptable LMs and broadening their applicability to other token-level tasks.

\centering{\faicon{github} \href{https://www.github.com/takelab/repetition-sl}{\texttt{takelab/repetition-sl}}}

\end{abstract}

\section{Introduction}
\begin{figure}[ht]
    \centering
    \includegraphics[width=.95\linewidth]{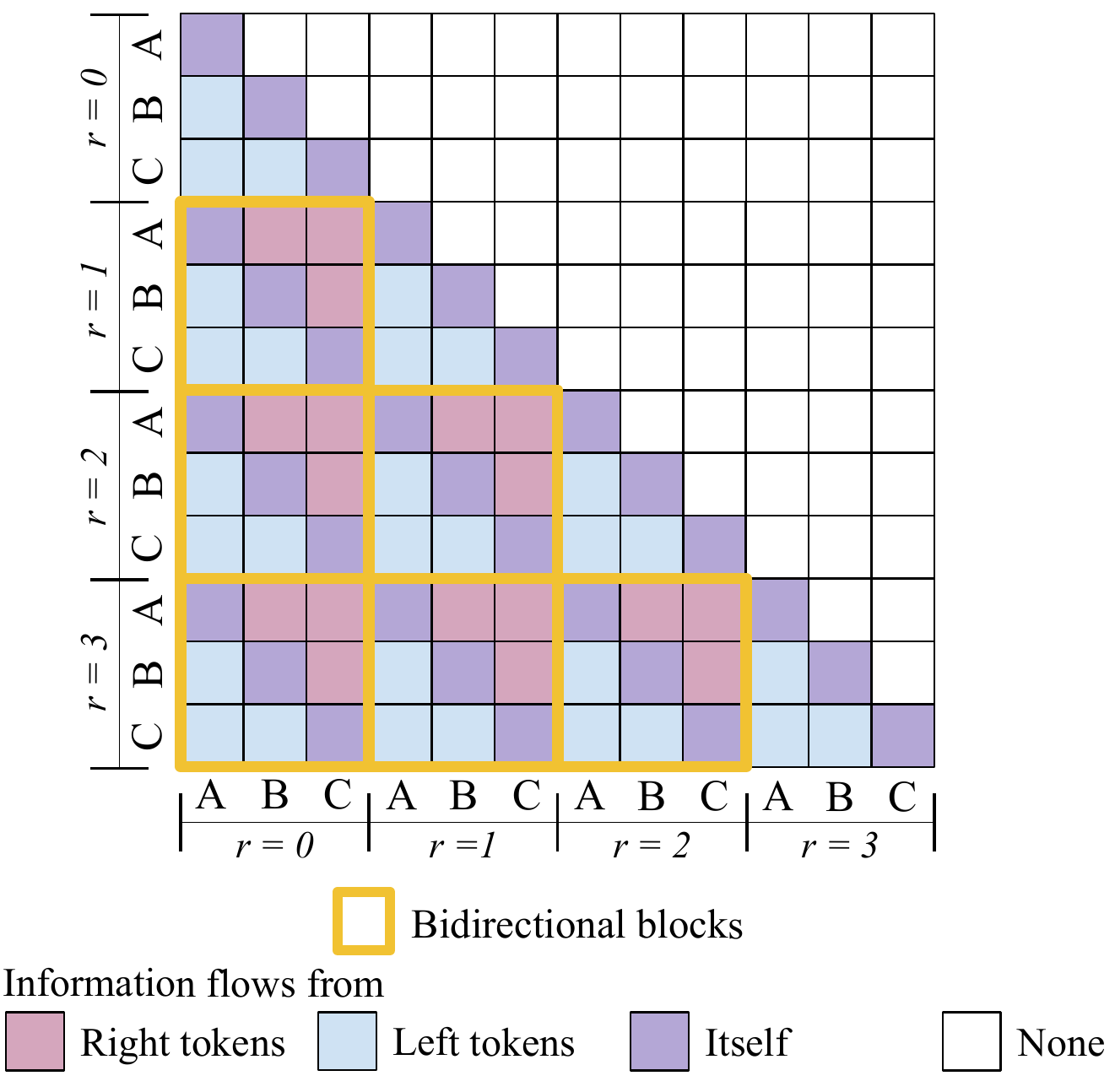}
    \caption{Attention weight matrix in a decoder block for an input sequence ``ABC''  repeated \textit{r} times. Sequence repetition (SR) forms bidirectional attention block regions with no modifications to the naturally unidirectional architecture of decoder-only language models. The share of bidirectional blocks contributing to the attention outputs increases with subsequent repetitions, enabling the model with a higher capacity to process tokens to the right of the currently processed token.}
    \label{fig:bidirectionality}
    \vspace{-1em}
\end{figure}

The NLP community has invested immense effort in training models for natural language understanding (NLU) tasks. Arguably, one of the more complex NLU tasks is sequence labeling (SL), where labels are assigned to individual tokens in a sequence. From named entity recognition and text chunking \citep{tjong-kim-sang-buchholz-2000-introduction, tjong-kim-sang-de-meulder-2003-introduction} to more challenging tasks such as event extraction \citep{doddington-etal-2004-automatic} and aspect-based sentiment analysis \citep{pontiki-etal-2014-semeval}, SL models rely on high-quality token embeddings. 

While encoder-only language models (LMs) such as BERT \citep{devlin-etal-2019-bert} and RoBERTa \citep{liu2019roberta} have long been the standard for SL tasks, scaling has enabled decoder-only LMs to produce superior embeddings, surpassing their encoder-only counterparts on various token- and sequence-level tasks \citep{behnamghader2024llm2vec,lin2025causal2vec}.
This trend has motivated research on decoder-as-encoder adaptation, typically by removing the causal mask (CM) in combination with parameter-efficient fine-tuning \citep{houlsby2019parameter,dukic-snajder-2024-looking}. 
While this modification provides the model with a fully bidirectional context, it is a substantial architectural change \citep{lin2025causal2vec}.
A parallel line of work has shown that simply repeating the input sequence can improve sequence-level representations in decoders \citep{xu-etal-2024-reading,springer2025repetition,duan2025retrieval}. Sequence repetition (SR) offers a less invasive method for incorporating bidirectionality without modifying the decoder’s core architecture, as illustrated in \autoref{fig:bidirectionality}. By repeating the input sequence several times, the decoder blocks form bidirectional regions inside attention matrices. However, despite its promise for sequence-level tasks,  the effect of SR on token-level embeddings---and, by extension, on SL performance in a decoder-as-encoder setup---remains largely unexplored.

To fill this gap, we explore the capacity of SR as a tool for token-level representation learning. Through controlled fine-tuning experiments, we show that SR implicitly enables bidirectionality within decoder-only models, yielding token-level embeddings that outperform both strong encoders and unmasked decoders across multiple SL benchmarks.
In a follow-up analysis, we find that, contrary to previous findings, repeating the input more than once often results in performance gains rather than degradation \citep{springer2025repetition}.\footnote{\footnotesize{Reported in experiments not included in the main paper \url{https://openreview.net/forum?id=Ahlrf2HGJR&noteId=zYNHEaQiWO}.}}
To mitigate the computational cost of sequence repetition, we utilize early exiting \citep{xin-etal-2020-deebert,10.5555/3495724.3497263,rep-etal-2024-electras} and extract embeddings from intermediate layers. Fine-tuning the model only up to a selected layer yields representations that perform on par with those from the final layer, while being at least 1.39$\times$ more efficient to compute.
These results reinforce findings that intermediate LM layers are critical for downstream tasks \citep{DBLP:journals/corr/abs-2403-02181, skean2025layer}.

The contributions of our work are threefold: (1) we find SR is a simple, practical, and effective strategy for SL with decoder-only models, outperforming more complex alternatives; (2) we show that, contrary to prior findings, performance improves as the number of sequence repetitions increases; (3) we demonstrate that early exiting mitigates the computational overhead of SR while maintaining performance competitive to last-layer embeddings.

\section{Sequence repetition}

Previous works have shown that repeating the input sequence once is a simple method for bypassing the CM used in decoder-only models
\citep{xu-etal-2024-reading, springer2025repetition}.
In this way, tokens in the repeated sequence have access to the entire sequence on the left-hand side, allowing the decoder-only model to function bidirectionally. However, while prior work has found that additional repetitions yield no further gains on sequence-level tasks \cite{xu-etal-2024-reading, springer2025repetition, duan2025retrieval}, our findings show that the opposite is true for token-level tasks, where performance improves with each repetition (\sect{sec:sr_res}). We hypothesize that this occurs because additional repetitions increase the model's effective processing capacity for each token. On the other hand, we posit that increasing the repetition count is not necessary for strong sequence-level performance because each token already encodes a compressed representation of the entire preceding sequence \cite{muennighoff2022sgpt}. We further note that fine-tuning is necessary for models to learn how to fully utilize this increased capacity, which is underexplored by previous works who studied the effect of repetition count increase only in the zero-shot setting.

We now outline our intuition of how repetitions enable bidirectional information flow during the computation of attention weights. 
Let $s = [t_1, \dots, t_n]$ be an input sequence of length $n$. We denote the repeated sequence as $s^{k}$, where $k=r+1$ is the total number of instances and $r$ is the number of repetitions. The final sequence has a length of $k \cdot n$, as illustrated in \autoref{fig:bidirectionality}. Let $A$ be the attention weight matrix computed via standard scaled dot-product attention \cite{vaswani2017attention} using causal masking \cite{radford2018improving}:
\begin{equation*}
    h_a(s) = A(s)V = \text{softmax}\left(\frac{QK^T + \mathit{CM}}{\sqrt{d}}\right)V,
\end{equation*} 
\noindent where $Q = s W_Q$, $K = s W_k$, and $V = sW_v$ are query, key, and value matrices, respectively, $d$ is the dimensionality of query and key vectors, and $\mathit{CM}$ is the causal mask defined as:
\[ \mathit{CM} =\begin{bmatrix}
    0 & -\infty & \cdots & -\infty \\
    0 & 0 & \cdots & -\infty \\
    \vdots & \vdots & \ddots & \vdots \\
    0 & 0 & \cdots & 0
\end{bmatrix}.
\] 
Due to causal masking, $A$ is a lower-triangular $n\times n$ matrix. For the repeated sequence $s^{k}$, which is composed of $k$ instances of $s$, the dimensions of $A$ grow to $kn \times kn$, and it can be viewed as a block matrix made of $k^2$ individual $n \times n$ submatrices: 
\[
A(s_k) = \begin{bmatrix}
    A_{11} & A_{12} & \cdots & A_{1k} \\
    A_{21} & A_{22} & \cdots & A_{2k} \\
    \vdots & \vdots & \ddots & \vdots \\
    A_{k1} & A_{k2} & \cdots & A_{kk}
\end{bmatrix}.
\] 
Each submatrix $A_{i,j}$ corresponds to the attention weights between tokens in the $i$-th and $j$-th instances of sequence $s$ within the repeated sequence $s^k$ (see \autoref{fig:bidirectionality}). Due to the causal mask, the structure of these submatrices is determined by their position: for $i > j$, the submatrix $A_{ij}$ is dense (full-rank); for $i = j$, $A_{ij}$ is lower-triangular; and for $i < j$, $A_{ij}$ is a zero matrix. 

In practice, this means that $A_{ij}$ contributes to the attention output with full bidirectional information for $i > j$ and with left and diagonal-only information for $i = j$, while zero matrices contribute nothing to the output. Thus, in the full attention matrix $A(s^k)$, there are $k(k+1)/2$ non-zero blocks that contribute any information, of which $k(k+1)/2 - k$ are fully bidirectional. The proportion of bidirectional blocks in $A(s^k)$ approaches $1$ as $k$ increases:
\begin{equation*}
\lim_{k\rightarrow\infty} \frac{k(k+1)/2-k}{k(k+1)/2} = \lim_{k\rightarrow\infty} \frac{k-1}{k+1} = 1
\end{equation*}

The representation of a token in the $k$-th instance of the sequence is computed using the attention weights in the blocks $A_{k1}\dots A_{kk}$. Since all of these blocks except the final one ($A_{kk}$) are bidirectional, the  overall attention mechanism more closely emulates full bidirectionality as $k$ increases. 

Naturally, this gain in processing capacity comes at a higher computational cost due to the $k$-fold increase in sequence length. However, we hypothesize that even small values of $k$ can yield substantial performance gains by closely approximating full bidirectionality. We verify this claim experimentally in subsequent sections.
\section{Experimental setup}

In this section, we first describe the standard suite of SL tasks we explore (\sect{sec:datasets}), the language models used in our experiments (\sect{sec:models}), and their optimization setup (\sect{sec:optimization}). We then outline the competitive baselines to which we compare our results (\sect{sec:baselines}). Finally, we detail the early-exit setup used to mitigate the computational cost of SR (\sect{sec:early_exit}).

\subsection{Datasets}
\label{sec:datasets}

\rparagraph{NER} 
For NER, we use the CoNLL03 dataset \citep{tjong-kim-sang-de-meulder-2003-introduction}, specifically the version available in HuggingFace datasets \citep{lhoest-etal-2021-datasets}, which follows the IOB2 sequence tagging scheme and provides predefined training, validation, and test splits.

\rparagraph{Aspect term extraction and polarity}
We use data from the restaurants domain of SemEval-2014 Task 4 \citep{pontiki-etal-2014-semeval} (Rest14). For our experiments, we merge the two main aspect-based sentiment analysis subtasks, aspect term extraction (ATE) and aspect term polarity (ATP), into a single SL task (ATE+ATP). We then tokenize the dataset with spaCy \citep{spacy} and match the given character spans of aspect terms with token spans to obtain IOB2 tags. While the training and test splits are predefined, we create our validation set by following prior work \citep{wang-etal-2021-automated} and randomly sampling 10\% of the training data.

\rparagraph{Slot labeling}
For this task, we use the NLU++ dataset \citep{casanueva-etal-2022-nlu} designed for NLU in task-oriented dialogue systems. The dataset is divided into two distinct domains (\textit{banking} and \textit{hotels}), each containing a mix of generic slots and domain-specific slots. To create a larger and more robust dataset for our experiments, we merge the two domains.
Merging domains results in a unified set of $17$ unique slot labels, as some generic slots overlap.
We create IOB2 tags for this combined dataset by mapping the character-level offsets of slot values to tokens generated by spaCy \citep{spacy}. Although the original authors recommend $k$-fold cross-validation, this approach is computationally prohibitive when fine-tuning large LMs. Therefore, we opt for fixed data splits, creating our training, validation, and test sets by shuffling and partitioning the data into a 70/10/20 ratio.

\rparagraph{Event trigger classification} 
The ACE05 dataset \citep{doddington-etal-2004-automatic} is a widely used event trigger classification (ETC) dataset. The ETC task combines two event extraction tasks into a single SL task: trigger identification, i.e., finding spans of tokens that evoke a particular event, and classifying them. We use the English training, validation, and test split obtained with the standard ACE pre-processing tool,\footnote{\footnotesize\url{https://bit.ly/ace2005-preprocessing}} which we also use to generate tokenized sentences and create IOB2 tags.

\subsection{Models}
\label{sec:models}
We conduct our experiments with seven open-weight decoder-only LMs from three different model families:
Gemma-7B \citep{team2024gemma}, Gemma2-2B, Gemma2-9B \citep{gemmateam2024gemma2improvingopen}, Qwen3-1.7B, Qwen3-4B, Qwen3-8B \citep{yang2025qwen3technicalreport}, and Mistral-7B \citep{jiang2023mistral}. 
By default, we use these decoder Transformers with a softmax token classification head on top of the model's Transformer body and train the models to output IOB2 tags. Using the same training and evaluation protocol, we compare the performance of decoder-only models to state-of-the-art encoder-only models: RoBERTa \citep{liu2019roberta} and ModernBERT \citep{warner-etal-2025-smarter}.
See \autoref{sec:setup_details} for implementation details.

\subsection{Optimization}
\label{sec:optimization}

We apply QLoRA \cite{dettmers2023qlora} to query, key, value, and output attention matrices in all decoder layers in our fine-tuning experiments (in early exit experiments with decoders we apply fine-tuning only up to a chosen intermediate layer). We use the following hyperparameters: fixed rank $r=16$, scaling parameter $\alpha=16$, dropout probability of $p=0.1$ and learning rate $2\mathrm{e}{-4}$. We train the models with the AdamW \citep{loshchilov2017decoupled} optimizer. The parameters of AdamW optimizer are left at default values when loaded from the PyTorch library $\beta_1=0.9, \beta_2=0.99, \epsilon=1\mathrm{e}{-8}$, with an exception of weight decay which is set to $0.05$. We employ gradient checkpointing with the reentrant implementation, using a batch size of $8$ and gradient accumulation over $4$ steps to fit our memory constraints, resulting in an effective batch size of $32$. All models were fine-tuned over $10$ epochs, and the best-performing weights were selected based on the highest micro F1 score on the validation set. For each fine-tuning experiment, we report the average micro F1 performance from five different runs (i.e., with different random seeds). For optimization details, please see \autoref{sec:experiment_details}.

\subsection{Baselines}
\label{sec:baselines}
We compare the performance of SR approaches to decoder-only baselines for SL listed below.

\rparagraph{Masked decoder} 
We fine-tune the base decoder model to output IOB2 tags with no modifications to the input sequence or the model's architecture. Under the SR formulation, this case corresponds to no repetitions ($r=0$).

\rparagraph{Unmasking} 
Following \cite{li2023label}, we remove CM from the attention computation at each layer, thereby allowing the models to utilize full bidirectional information in the input sequence at each token, effectively transforming them to encoders.

\rparagraph{Layer-group unmasking} 
\citet{dukic-snajder-2024-looking} showed that, for optimal SL performance, CM should be removed only from selected layer groups, most commonly from middle layers. This is sensible, since intermediate layers have been shown to be important for improvements on downstream tasks \citep{skean2025layer}. For a model with $N$ layers, we remove the CM from the middle $N/3$ layers. See \autoref{app:unmasking_middle_imp} for details on how we select the exact indices to perform unmasking.
\begin{table*}[ht]
\centering
\small
\renewcommand{\arraystretch}{1.2}
\begin{tabular*}{\textwidth}{l@{\extracolsep{\fill}}lccccc}
\toprule
\textbf{Model} & \textbf{Method} & \textbf{NLU++} & \textbf{ACE05} & \textbf{Rest14} & \textbf{CoNLL03} & \textbf{Average} \\
\midrule\multirow{3}{*}{\textbf{Gemma-7B}} & Sequence repetition ($r = 4$)& $\underline{\textbf{81.29}}_{{\pm 2.23}}$ & $\underline{74.27}_{{\pm 1.41}}$ & $82.40_{{\pm 0.80}}$ & $\underline{93.38}_{{\pm 0.18}}$ & $\underline{82.84}_{{\pm 0.69}}$ \\& Middle unmasking & $79.90_{\pm 3.36}$ & $72.15_{\pm 5.54}$ & $\underline{\textbf{83.64}}_{\pm 0.68}$ & $91.46_{\pm 4.93}$ & $81.79_{\pm 2.04}$ \\& Full unmasking & $74.34_{\pm 2.77}$ & $73.08_{\pm 6.61}$ & $76.63_{\pm 4.97}$ & $92.75_{\pm 0.16}$ & $79.20_{\pm 2.18}$ \\\midrule\multirow{3}{*}{\textbf{Gemma2-2B}} & Sequence repetition ($r = 2$)& $\underline{79.98}_{\pm 1.21}^{}$ & $72.86_{\pm 3.99}^{}$ & $\underline{83.03}_{\pm 1.03}^{}$ & $\underline{93.05}_{\pm 0.29}^{}$ & $\underline{82.23}_{\pm 1.00}$ \\& Middle unmasking & $78.21_{\pm 2.03}$ & $72.30_{\pm 3.99}$ & $82.21_{\pm 0.40}$ & $92.60_{\pm 0.32}$ & $81.33_{\pm 1.13}$ \\& Full unmasking & $76.42_{\pm 2.16}$ & $\underline{74.33}_{\pm 4.20}$ & $81.26_{\pm 0.64}$ & $92.82_{\pm 0.36}$ & $81.21_{\pm 1.19}$ \\\midrule\multirow{3}{*}{\textbf{Gemma2-9B}} & Sequence repetition ($r = 2$)& $\underline{78.83}_{\pm 2.47}^{}$ & $74.49_{\pm 2.93}^{}$ & $\underline{83.51}_{\pm 0.89}^{}$ & $\underline{93.58}_{\pm 0.24}^{}$ & $\underline{82.60}_{\pm 0.79}$ \\& Middle unmasking & $77.63_{\pm 1.23}$ & $\underline{75.91}_{\pm 2.93}$ & $83.34_{\pm 0.97}$ & $93.17_{\pm 0.36}$ & $82.51_{\pm 0.84}$ \\& Full unmasking & $76.83_{\pm 2.09}$ & $75.79_{\pm 3.70}$ & $82.55_{\pm 1.28}$ & $93.44_{\pm 0.13}$ & $82.15_{\pm 1.11}$ \\\midrule\multirow{3}{*}{\textbf{Mistral-7B}} & Sequence repetition ($r = 4$)& $\underline{80.28}_{\pm 1.78}^{}$ & $\underline{\textbf{77.79}}_{\pm 3.31}^{}$ & $\underline{83.31}_{\pm 1.57}^{}$ & $\underline{\textbf{93.79}}_{\pm 0.19}^{}$ & $\underline{\textbf{83.79}}_{\pm 1.59}$ \\& Middle unmasking & $79.35_{\pm 0.52}$ & $77.56_{\pm 3.31}$ & $83.00_{\pm 0.89}$ & $93.48_{\pm 0.12}$ & $83.35_{\pm 0.87}$ \\& Full unmasking & $80.23_{\pm 1.24}$ & $76.26_{\pm 3.77}$ & $83.18_{\pm 0.84}$ & $93.69_{\pm 0.24}$ & $83.34_{\pm 1.02}$ \\\midrule\multirow{3}{*}{\textbf{Qwen3-1.7B}} & Sequence repetition ($r = 4$)& $\underline{77.75}_{\pm 1.43}^{}$ & $\underline{70.18}_{\pm 3.87}^{}$ & $\underline{79.26}_{\pm 1.03}^{}$ & $\underline{91.56}_{\pm 0.41}^{}$ & $\underline{79.69}_{\pm 0.97}$ \\& Middle unmasking & $73.67_{\pm 1.46}$ & $66.71_{\pm 3.87}$ & $75.71_{\pm 1.02}$ & $90.31_{\pm 0.42}$ & $76.60_{\pm 1.07}$ \\& Full unmasking & $67.28_{\pm 4.24}$ & $68.68_{\pm 3.39}$ & $75.32_{\pm 1.99}$ & $90.27_{\pm 0.45}$ & $75.39_{\pm 1.45}$ \\\midrule\multirow{3}{*}{\textbf{Qwen3-4B}} & Sequence repetition ($r = 8$)& $\underline{77.97}_{\pm 1.29}^{}$ & $\underline{72.58}_{\pm 2.91}^{}$ & $\underline{81.22}_{\pm 0.48}^{}$ & $\underline{92.33}_{\pm 0.28}^{}$ & $\underline{81.02}_{\pm 1.35}$ \\& Middle unmasking & $74.56_{\pm 0.81}$ & $69.73_{\pm 2.91}$ & $79.60_{\pm 0.41}$ & $91.87_{\pm 0.34}$ & $78.94_{\pm 0.77}$ \\& Full unmasking & $73.79_{\pm 1.17}$ & $70.59_{\pm 4.48}$ & $77.76_{\pm 0.62}$ & $91.17_{\pm 0.17}$ & $78.33_{\pm 1.17}$ \\\midrule\multirow{3}{*}{\textbf{Qwen3-8B}} & Sequence repetition ($r = 4$)& $\underline{77.61}_{\pm 2.52}^{}$ & $\underline{72.11}_{\pm 2.97}^{}$ & $\underline{82.27}_{\pm 1.11}^{}$ & $\underline{92.74}_{\pm 0.26}^{}$ & $\underline{81.18}_{\pm 1.62}$ \\& Middle unmasking & $75.06_{\pm 1.18}$ & $71.45_{\pm 2.97}$ & $80.17_{\pm 0.74}$ & $92.23_{\pm 0.38}$ & $79.73_{\pm 0.83}$ \\& Full unmasking & $74.45_{\pm 2.72}$ & $69.01_{\pm 3.11}$ & $80.01_{\pm 0.42}$ & $91.99_{\pm 0.29}$ & $78.86_{\pm 1.04}$ \\\midrule{\textbf{ModernBERT}} & - & $70.50_{\pm 3.24}$ & $68.30_{\pm 4.18}$ & $75.51_{\pm 1.01}$ & $90.02_{\pm 0.23}$ & $76.08_{\pm 1.35}$ \\{\textbf{RoBERTa}} & - & $72.22_{\pm 1.68}$ & $68.96_{\pm 8.04}$ & $80.30_{\pm 0.78}$ & $92.64_{\pm 0.12}$ & $78.53_{\pm 2.06}$ \\\bottomrule
\end{tabular*}
\caption{Micro F1 scores (mean $\pm 95$\% confidence interval) across Rest14, ACE05, CoNLL03, and NLU++ test sets. \textbf{Bold} values indicate the highest mean per dataset within each group. Values obtained with the best-performing strategy for a given model and a dataset are \underline{underlined}. All results are averages of five runs. We report the results for $r$ which overall had the best average F1 score for a given model.}
\label{tab:main_res}
\end{table*}

\subsection{Early exit} 
\label{sec:early_exit}
As SR introduces additional input processing, we propose combining it with early exit for a trade-off between quality and efficiency. Following prior findings that earlier layers can yield embeddings comparable in quality to those from later layers for downstream tasks \citep{xin-etal-2020-deebert,10.5555/3495724.3497263,rep-etal-2024-electras}, we investigate fine-tuning the model up to, but not including, layer $L$ in conjunction with SR. Specifically, we consider a model with $N=32$ layers, and fine-tune up to layer $L \in \{9,14,19,24\}$. This improves computational efficiency by factors ranging from $4\times$ to $1.39\times$ when exiting at layers $9$ and $24$, respectively.

\section{Results}
\label{sec:results}

In this section, we first compare SR to encoder-only models and decoder-only approaches to SL (\sect{sec:main_results}). We then analyze SR performance as the number of repetitions is varied (\sect{sec:sr_res}). We follow up by exploring early exiting as a strategy of mitigating additional compute incurred by SR (\sect{sec:early_exit_res}), comparing early exiting decoders with encoders (\sect{sec:juxtaposing_dec_enc}), and profiling how early exiting and SR affect inference speed (\sect{sec:inference_profiling}). Finally, we conduct a qualitative analysis of performance between adaptation strategies (\sect{sec:qualitative}).
\begin{figure*}[ht]
    \centering
    \includegraphics[width=1.0\linewidth]{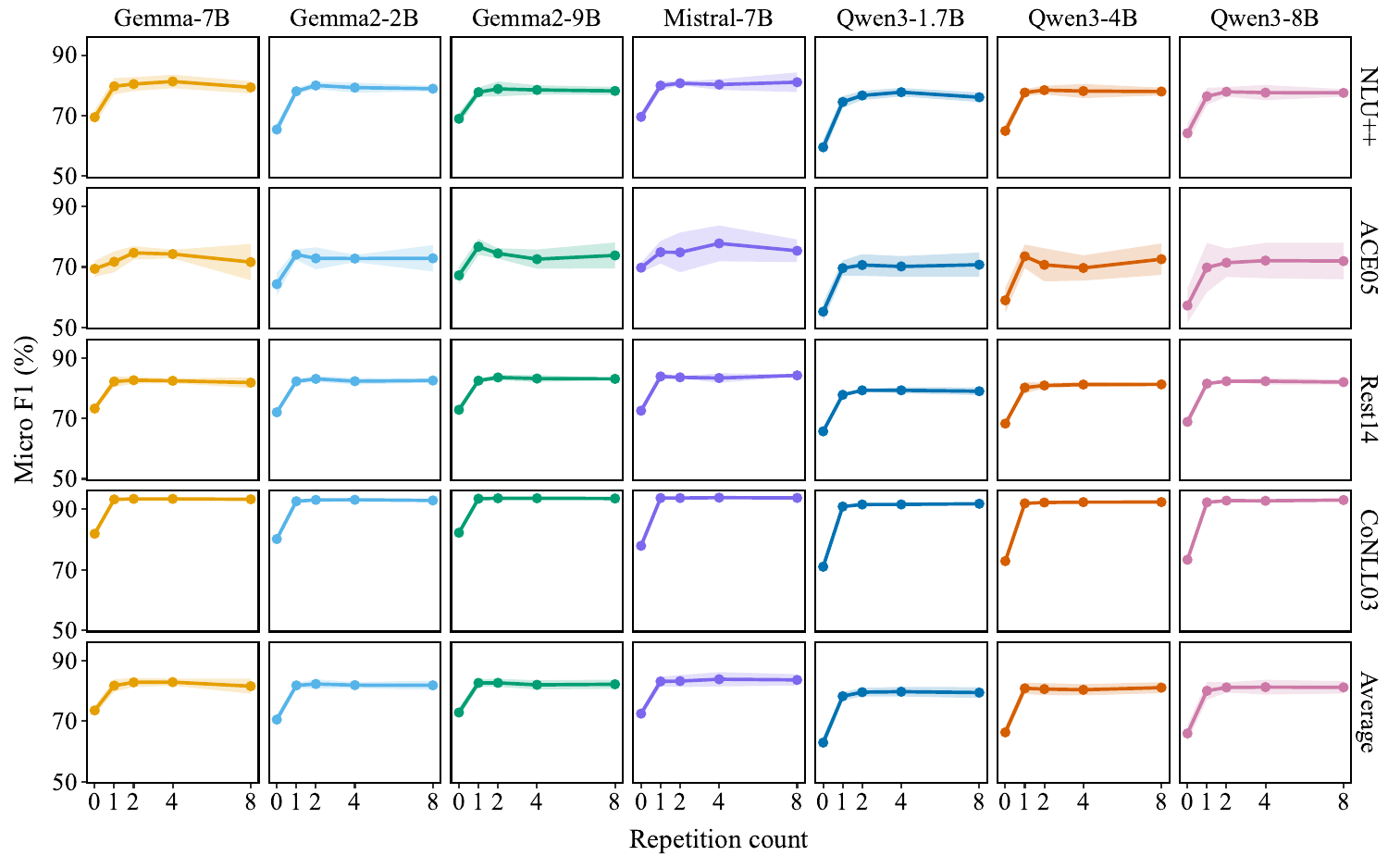}
    \caption{Micro F1 performance for SR across models and test splits of each dataset. Subplots are aligned in a 5x7 grid, where rows and columns correspond to datasets and models, respectively. In the fifth row, we display the performance averaged over all the datasets. Shaded bands denote $\pm95\%$ CI. All results are averages of five runs.}
    \label{fig:repeat_k}
\end{figure*}
\subsection{Performance on sequence labeling}
\label{sec:main_results}

\begin{figure*}[ht]
    \centering
    \includegraphics[width=\linewidth]{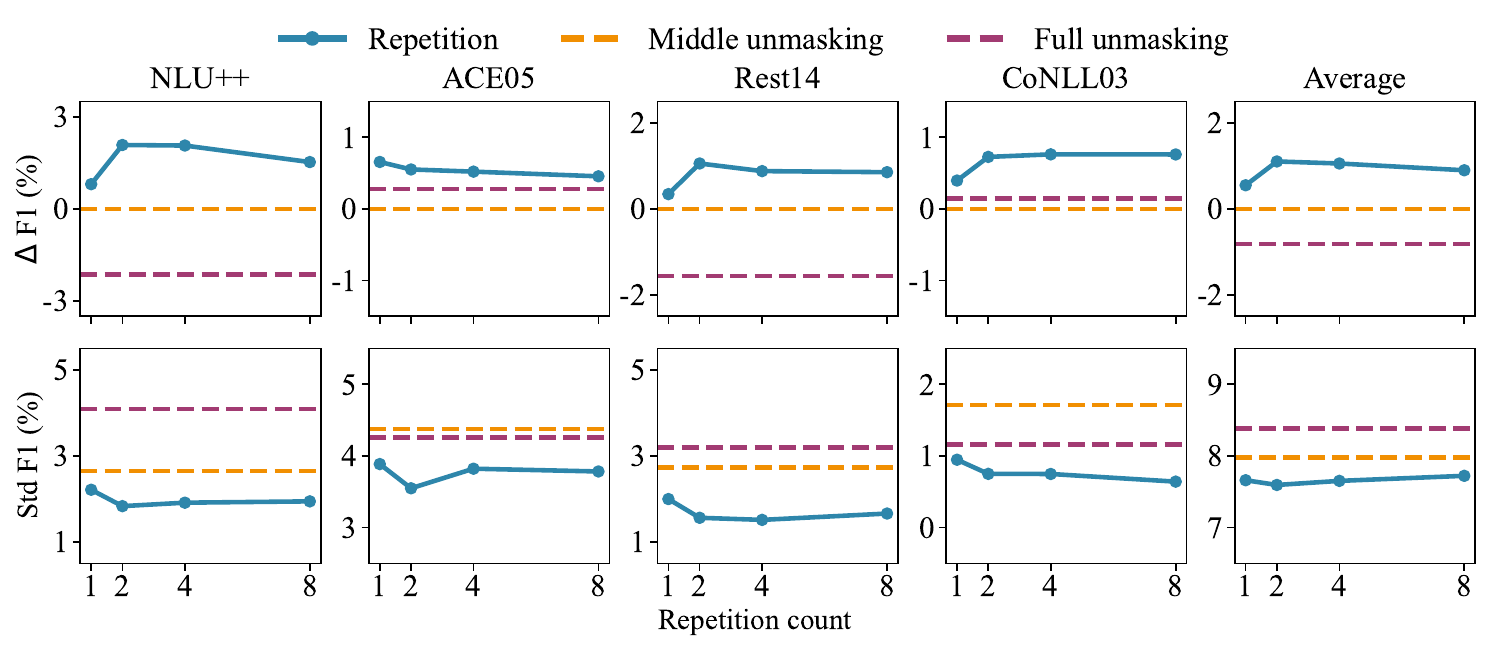}
    \caption{F1 score gains over middle unmasking (higher is better) and their standard deviations (lower is better) computed over all decoder-only models for each dataset and adaptation strategy.}
    \label{fig:agg-results}
\end{figure*}

We report results of the best-performing encoder-based methods, unmasking baselines, and SR on the test splits of datasets in \autoref{tab:main_res}.
Generally, decoder-based models outperform encoders. Specifically, encoders lag behind unmasked decoders and even more so behind SR decoders, which outperform them by a large margin.
When considering unmasking baselines, unmasking only the middle layers generally outperforms removing the CM entirely. On Rest14, Gemma-7B achieves the best results among all models with middle-layer unmasking. Other models also benefit more from partial bidirectionality than from full bidirectionality. The exception is Mistral-7B, where full unmasking surpasses middle-layer unmasking on all SL datasets except ACE05. This suggests that Mistral-7B benefits more from full bidirectionality than solely from the bidirectionality of the middle layers. 
Most importantly, the best overall results are achieved with SR models, where Mistral-7B stands out as the best-performing model, achieving the highest average score and outperforming other models on CoNLL03 and ACE05 for $r=4$. This trend is even more apparent in \autoref{fig:repeat_k} and \autoref{fig:repeat_k_valid}, where Mistral-7B leverages additional repetitions most effectively across models and datasets on both the test and validation sets, respectively. Finally, in \autoref{fig:agg-results}, we present average differences in performance of SR and full unmasking against middle unmasking. We observe clear gains over unmasking strategies both in terms of better performance (higher mean F1 score) and higher stability (lower standard deviation of F1 scores).

\subsection{Effect of multiple sequence repetitions}

\label{sec:sr_res}
We now investigate how model performance behaves when varying the number of repetitions.
We report the results of SR fine-tuning for $r \in \{0, 1, 2, 4, 8\}$ in \autoref{fig:repeat_k}.
Repeating the sequence once ($r=1$) yields clear gains over the masked decoder ($r=0$).
This consistency is observed across all SL datasets, models, and both validation and test sets. The highest jump in performance is from $r=0$ to $r=1$.
Increasing the number of sequence repetitions further improves the performance for most models and datasets. As shown in \autoref{fig:agg-results}, setting $r=2$ improves the average performance of the models and reduces the variance in the results. Performance gains generally saturate at $r=2$ or $r=4$ and remain stable with additional repetitions, contrary to prior reports of degradation on sequence-level tasks for $r>1$ \citep{springer2025repetition}.
Mistral-7B stands out as the best-performing model, especially as the number of repetitions approaches $r=8$.
Smaller models (Qwen3-1.7B and Gemma2-2B) exhibit a noticeable performance drop in most cases after $r=2$ or $r=4$, suggesting that they lack the capacity to effectively utilize additional repetitions.
However, already at $4$B parameters (Qwen3-4B), performance stabilizes and saturates as $r$ grows. We hypothesize that larger models, with larger effective context windows, could leverage additional repetitions to a greater extent. 
The relationship between micro F1 scores and repetition counts is mostly consistent across models and datasets. However, we observe anomalies on ACE05, which we attribute to differences in class distribution in training and test datasets \cite{wang-etal-2020-maven,dukic-snajder-2024-looking}. We hypothesize that because higher repetition counts increase the effective processing capacity of the models, they are more prone to overfitting on the training set, resulting in a higher variance in performance. This is further corroborated by slightly larger confidence intervals in F1 scores of larger models ($\geq$7B parameters) as opposed to smaller ones ($\leq$2B parameters). Note that we did not tune hyperparameters for each dataset specifically. Therefore, this instability can likely be resolved through a more careful hyperparameter selection, which we leave for future work.

\subsection{Qualitative analysis of adaptation strategy performance}
\label{sec:qualitative}
\begin{figure}[ht]
    \centering
    \includegraphics[width=\linewidth]{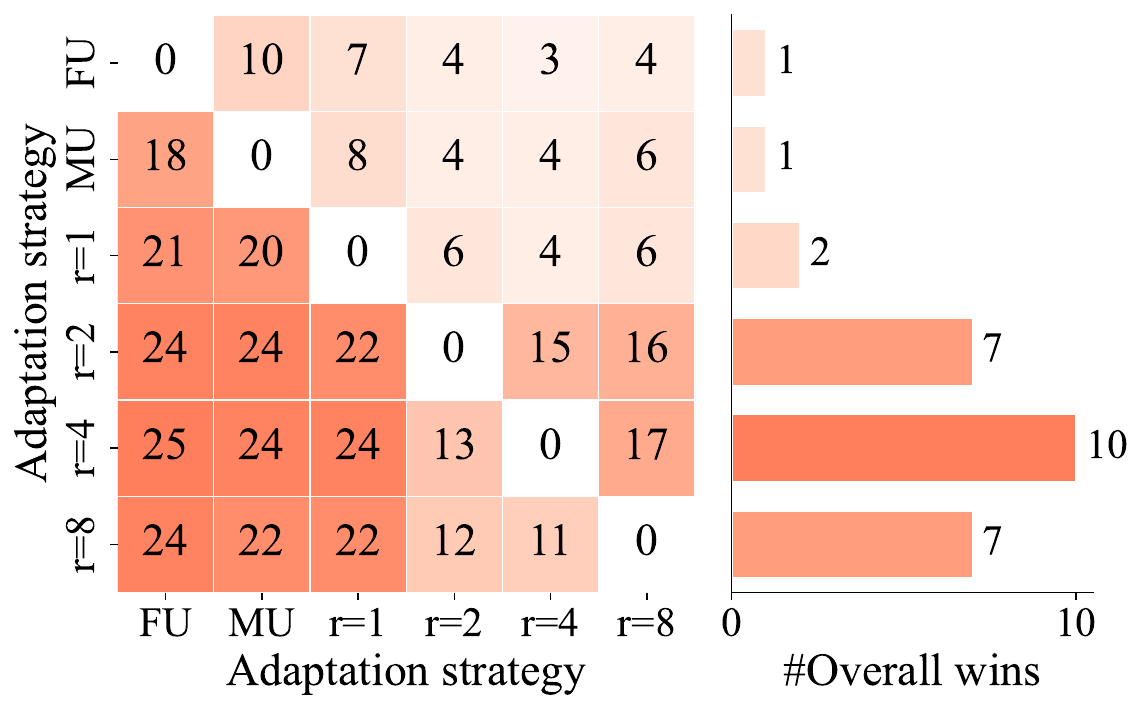}
    \caption{The heatmap on the left-hand side is an adaptation strategy pairwise superiority matrix. Cell $i,j$ contains the number of (model, dataset) pairs for which an adaptation strategy at row $i$ achieved a superior performance to the strategy at column $j$. On the right-hand side, we show the number of (model, dataset) pairs (28 in total) for which an adaptation strategy in the corresponding row of the heatmap achieved performance superior to all other strategies. FU and MU stand for full and middle unmasking, respectively.}
    \label{fig:qualitative}
\end{figure}

\autoref{fig:qualitative} summarizes our performance comparisons, based on the full results presented in \autoref{app:full_results}. The figure shows both pairwise results for each strategy and the total number of wins per strategy across all (model, dataset) pairs.

SR with more than one repetition ($r > 1$) proves to be the superior strategy, winning in 24 out of 28 (model, dataset) pairs, thus demonstrating the efficacy of multiple repetitions. Although overall performance tends to saturate around $r=4$ (cf. \autoref{fig:repeat_k}), this high win rate is maintained up to $r=8$.  Full and middle unmasking achieve superior performance for only two (model, dataset) pairs: (Gemma-7B, Rest14) and (Gemma2-2B, ACE05), respectively, while $r=1$ was the winner for (Gemma2-9B, ACE05), and (Mistral-7B, ACE05), which is consistent with generally anomalous behavior on ACE05. 

\begin{figure*}[ht]
    \centering
    \includegraphics[width=\linewidth]{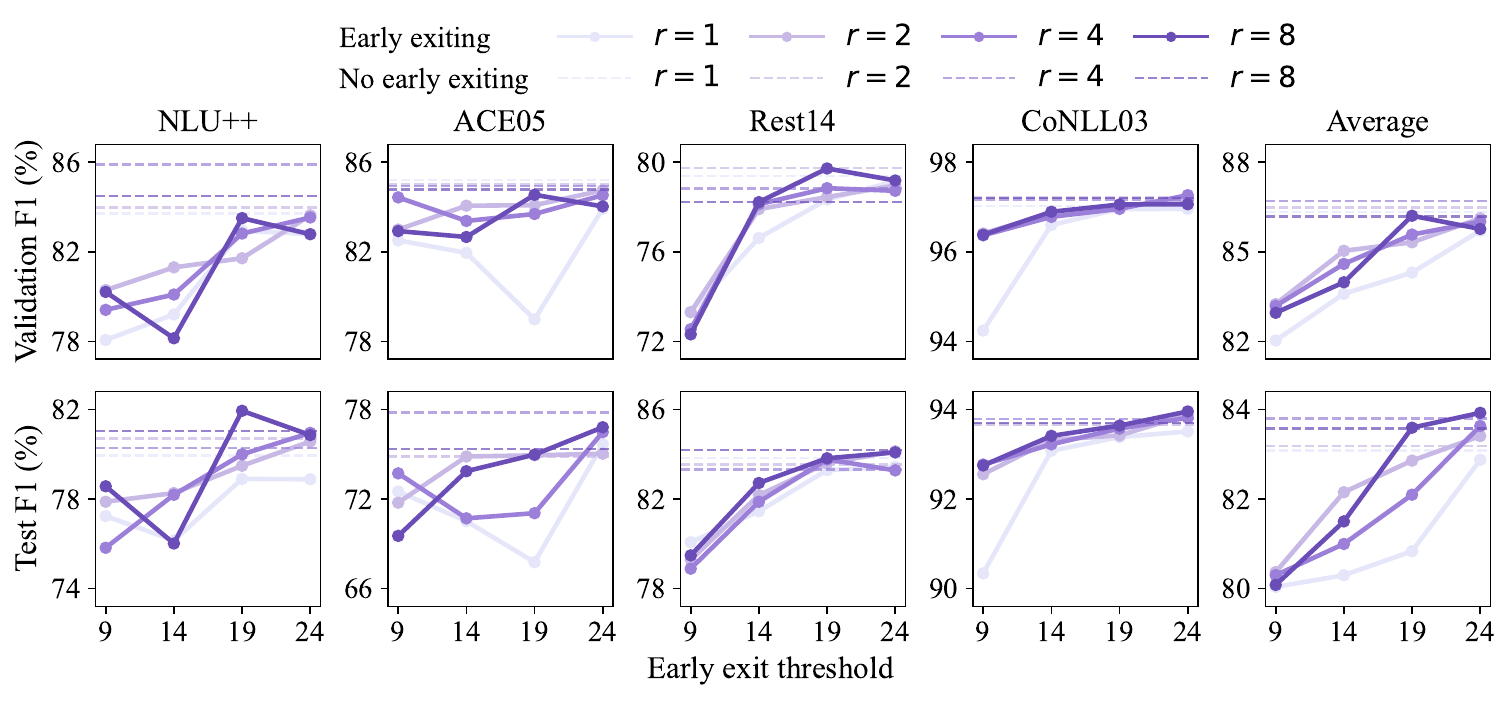}
    \caption{Micro F1 scores for early exit experiments across datasets and splits for Mistral-7B model. Each column corresponds to a dataset, with the top row reserved for validation and the bottom row for testing. Solid lines trace mean F1 over exit layers ($9$–$24$ out of $32$) for sequence repetition  $r$, while dashed lines denote the performance of the model without early exiting (exit@$L=32$). All results are averages of five runs.}
    \label{fig:early_exit}
    \vspace{-1em}
\end{figure*}

\subsection{Improving efficiency by early exiting}
\label{sec:early_exit_res}

SR yields higher-quality token-level embeddings than unmasking, leading to greater gains on SL tasks. However, this method introduces a significant computational cost, especially as the number of repetitions increases. To mitigate this overhead, we explore an early-exit strategy.
We hypothesize that intermediate-layer embeddings can yield competitive performance compared to last-layer embeddings.
To verify this, we fine-tuned our best-performing decoder (Mistral-7B) with sequence repetitions $r\in\{1,2,4,8\}$ up to, but not including, layer $L\in\{9,14,19,24\}$.
Mistral-7B has $32$ layers, and exiting at layer $L=24$ results in a roughly $28\%$ reduction in the number of parameters. 
We report results on validation and test sets in \autoref{fig:early_exit}. \textit{Earliest} layer embeddings from $L=9$ do not perform on par with the embeddings from the last layer, especially in the case when $r=1$. However, embeddings from layers as early as $L=14$ can already match the performance of the last layer (see results on Rest14).
Generally, the best-performing embeddings are obtained from early exits at layers $L=19$ and $L=24$, and their quality improves with more repetitions.
In some cases, these intermediate embeddings can even outperform those from the last layer (see Rest14 validation and NLU++ test for $L=19$), indicating that intermediate layer embeddings might be more suitable for some downstream tasks, a finding that aligns with previous work \citep{skean2025layer}.

\subsection{Inference profiling for early exiting} \label{sec:inference_profiling}

\autoref{tab:profiling_results} provides profiling results of inference speedup with respect to repetition and early exit factors. Early exiting significantly increases inference speed. For instance, at $r=2$ and exit at $L=24$, the model matches the base model's inference time, demonstrating that early exiting effectively mitigates the additional computational overhead, while performing on par with the base model (see \autoref{fig:early_exit}).
\begin{table}[ht]
\centering
\resizebox{\linewidth}{!}{
\begin{tabular}{lcccc}
\toprule
\textbf{Configuration} & $r=1$ & $r=2$ & $r=4$ & $r=8$ \\
\midrule
exit@$L=9$ & \cellcolor{steelblue!74}{$3.96$} & \cellcolor{steelblue!47}{$2.89$} & \cellcolor{steelblue!28}{$2.13$} & \cellcolor{steelblue!12}{$1.47$}\\
exit@$L=14$ & \cellcolor{steelblue!29}{$2.15$} & \cellcolor{steelblue!18}{$1.74$} & \cellcolor{steelblue!6}{$1.25$} & \cellcolor{burntorange!5}{$0.82$}\\
exit@$L=19$ & \cellcolor{steelblue!13}{$1.51$} & \cellcolor{steelblue!6}{$1.25$} & \cellcolor{burntorange!1}{$0.98$} & \cellcolor{burntorange!17}{$0.59$}\\
exit@$L=24$ & \cellcolor{steelblue!4}{$1.17$} & \cellcolor{burntorange!1}{$0.98$} & \cellcolor{burntorange!10}{$0.72$} & \cellcolor{burntorange!29}{$0.46$}\\
\bottomrule
\end{tabular}}
\caption{Profiling results calculated on the validation sets of NLU++, ACE05, Rest14, and CoNLL2003. We report speedup factors relative to the base Mistral-7B (no early exiting, $r=0$) across different early exiting configurations and multiple repetitions ($r$).}
\label{tab:profiling_results}
\vspace{-1em}
\end{table}

\subsection{Juxtaposing decoders and encoders of comparable size} 

\label{sec:juxtaposing_dec_enc}
\begin{figure*}
    \centering
    \includegraphics[width=\linewidth]{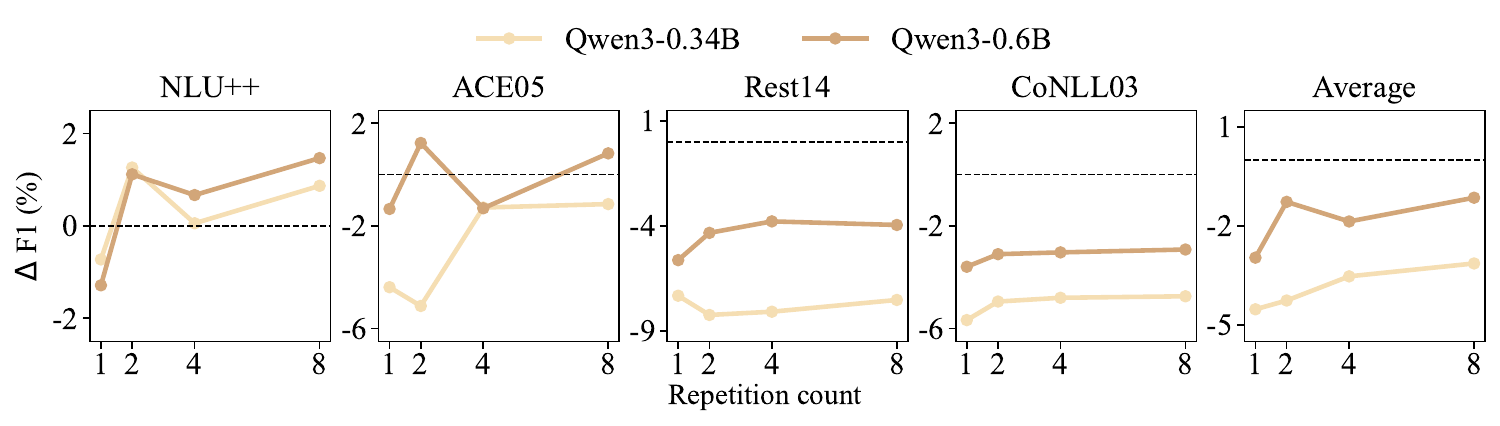}
    \caption{Comparison of Qwen3-0.6B (full model) and Qwen3-0.34B (early exit@$L=13$) against RoBERTa (0.35B). We report the absolute differences ($\Delta F_1 =$ Qwen $F_1$ - RoBERTa $F_1$) averaged over five seeds.}
    \label{fig:qwen_vs_roberta}
\end{figure*}

\autoref{tab:main_res} juxtaposes decoders with strong encoders, but not on the same scale. For example, we compared RoBERTa with 0.35B parameters to decoders from 1.7B to 9B parameters. To account for the model size factor, we level the playing field and experiment with much smaller Qwen3 decoders with 0.34B and 0.6B parameters. We obtain the Qwen3-0.34B model by early exiting at $L=13$ of Qwen3-0.6B. We fine-tune the models with SR for $r\in \{1,2,4,8 \}$ and report the absolute differences from RoBERTa in \autoref{fig:qwen_vs_roberta}. The results show that even the smallest decoder-only model can perform on par with the best encoder-only model when using our SR method, with higher $r$ generally yielding improvements. The improved performance of decoders over encoders at the billion-parameter scale, combined with SR, justifies the computational cost.

\section{Related work}

\rparagraph{Decoder-as-encoder adaptation}
Causal LM pre-training enforces a causal mask that blocks attention to future tokens, a property that benefits left-to-right generation but prohibits individual token-level representations from being fully contextualized. Despite this architectural constraint, decoder-only LMs yield strong sequence embeddings without architectural modifications via mean pooling or last token extraction \cite{muennighoff2022sgpt}. To further enhance embedding performance, recent works enable access to bidirectional information through architectural interventions. \citet{behnamghader2024llm2vec} propose LLM2Vec, a method consisting of three interventions on decoders: causal mask removal, self-supervised training for masked next token prediction, and unsupervised learning to improve sentence embedding capabilities of the model. Similarly, NV-Embed \cite{lee2025nvembed} also remove the causal mask and add a trainable latent attention layer to pool the sequence outputs into a more expressive fixed-size embedding, while Causal2Vec \cite{lin2025causal2vec} utilize an encoder model to produce a sentence embedding aware of bidirectional context termed \textit{Contextual token}, which they prepend to the decoder input and produce the final sentence embedding.

Causal mask removal was also shown to be an effective strategy for token-level tasks such as SL \cite{li2023label, dukic-snajder-2024-looking}. In a similar vein, \citet{huang2025transforming} improve question-answering performance of decoders by aggregating outputs from standard causal attention and right-only attention obtained via a mask complementary to the causal mask.

In contrast to the aforementioned methods, which require extenstive training for decoder-as-encoder adaptation, \citet{springer2025repetition} show that bidirectionality can be enabled without architectural modifications by concatenating a sequence to itself. Other approaches improve representations through prompting \cite{jiang-etal-2024-scaling, lei-etal-2024-meta}, iterative token embedding injection \cite{fu-etal-2025-token}, re-routing key-value states \cite{tang2026kv}, or training-free architectural reconfiguration \cite{lin-etal-2025-look}. We continue the line of work investigating bidirectionality without architectural modifications for token-level tasks.

\rparagraph{Sequence repetition}
In addition to \citet{springer2025repetition}, who demonstrated that sequence repetition can strengthen sentence representations, related works explore re-reading prompts to improve text understanding in LMs \citep{xu-etal-2024-reading, leviathan2025prompt}, and analyze mechanisms of repetition in LMs \cite{mahaut2025repetitions, hiraoka-inui-2025-repetition}. Our work complements this line of research by investigating the effects of repetition on token-level embeddings and, consequently, on downstream SL performance.

\rparagraph{Intermediate-layer embeddings quality}
Early exiting techniques utilize intermediate hidden representations of transformer-based models, rather than final-layer ones, motivated by the fact that the residual stream ties semantic spaces between layers \citep{geva-etal-2022-transformer}.
Early exiting offers a simple and effective approach for balancing inference costs by leveraging the computational nature of transformers \citep{teerapittayanon2016branchynet, huang2018multiscale, jazbec2024fast}.
Indeed, extracting embeddings from intermediate layers has been shown to yield superior sequence-level representations compared to final-layer pooling \citep{rep-etal-2024-electras}. This benefit has been observed across numerous downstream tasks, not only in NLP but also in computer vision \citep{alain2017understanding,skean2025layer}. Motivated by these insights, we explore the viability of early exiting as a means to counteract the computational overhead of SR.

\section{Conclusion}

While powerful decoder-only models are achieving competitive results on sequence labeling (SL) tasks, adapting their unidirectional architecture to inherently bidirectional tasks often requires invasive changes, such as removing the causal mask.
In contrast, sequence repetition (SR) is a more straightforward, input-level method that enables bidirectionality without altering the model's core structure. We show that SR improves token-level representations in decoder-as-encoder setups, yielding consistent SL gains over strong encoders and unmasked decoders.
Moreover, we demonstrate that performance continues to improve with multiple repetitions and that an early-exit strategy utilizing intermediate-layer embeddings offers a favorable quality–efficiency trade-off. 
A qualitative analysis contrasting model performance across adaptation strategies further highlight performance benefits obtained by SR, highlighting the effectiveness of this simple method. %
Our results establish SR as an effective, architecture-preserving method for achieving bidirectionality in decoder-only LMs. Future work could extend this approach to multilingual and structured prediction settings and explore adaptive repetition schedules conditioned on instance complexity.

%
%
\section*{Limitations}
Our approach has a number of key limitations.
Firstly, we report means over five runs with different random seeds to improve reliability. Increasing the number of runs or utilizing $k$-fold cross-validation would further improve  robustness of our findings.
Next, we maintain fixed learning rates and hyperparameters across experiments, may have led to suboptimal adaptation for SL and underutilization of the increased processing capacity as a result of higher repetition counts.
Furthermore, additional unmasking configurations other than unmasking only the middle layers would likely reduce the performance gap between SR and unmasking baselines. However, trying out even a subset of all possible layer group unmasking configurations would be immensely computationally demanding. 
While the open-weight landscape is rich, our experiments target models with $1.7$ to $9$ billion parameters.
Experimenting with larger open-weight LLMs could yield further insights into the behavior of decoders when presented with additional sequence repetitions.
Crucially, our experiments are limited to English language datasets. Extending to other languages and more diverse corpora should be explored as a next step. Additionally, we conducted our experiments on four SL datasets, which, although diverse, may not capture all the peculiarities of the broader landscape of SL tasks.
Finally, although sequence repetitions evidently help to improve the performance of decoders on SL tasks, they become increasingly inefficient to compute beyond $r=4$, even when early exiting at layer $L=24$, offering diminishing returns relative to their additional cost.
\section*{Acknowledgements}
Marko Čuljak is supported by the Croatian Science Foundation (HRZZ) Young Researchers’ Career Development Project (grant DOK-NPOO-2023-10-1392). The experiments were conducted on the HPC infrastructure maintained by University of Zagreb University Computing Centre (Srce). We thank Ana Barić and Laura Majer for providing additional computational resources when needed.

\bibliography{custom, anthology}

\appendix

\section{Details of experimental setup}
\label{sec:setup_details}

\subsection{Models and datasets}
In \autoref{tab:models}, we show the list of models we used and their HuggingFace identifiers. In \autoref{tab:dataset_stats}, we present the statistics of the datasets on which we conduct our experiments.

\subsection{Classification with decoders}
\label{subsec:decoder_as_encoder}

To perform classification tasks, decoder models must be equipped with a classification layer.
We implement this by using a softmax linear layer whose input dimension is the size of the hidden vectors produced by the model, and output dimension of the number of target classes. This assigns a set of logits to every token in the input sequence for each possible label. We input the final hidden state after the decoder model's forward pass into this layer.

\subsection{LLM adaptation details}
\label{subsec:implementation_details}

In this section, we outline the methodology behind modifications to decoder-only models: middle layer unmasking (\sect{app:unmasking_middle_imp}) and sequence repetition (\sect{app:seq_rep}).

\subsubsection{Unmasking of middle layers} 
\label{app:unmasking_middle_imp}

Firstly, we determine which layers are the middle ones in a given model.
The selection of these layers is determined dynamically based on the total number of hidden layers, denoted as $N = \texttt{config.num\_hidden\_layers}$. The interval bounds are computed as follows:
\[
N_u = \left\lfloor \frac{N}{3} \right\rfloor_{\text{even}}
\]
\[
lb = \frac{N}{2} - 1 - \frac{N_u}{2}, \quad
ub = \frac{N}{2} + \frac{N_u}{2}.
\]
This formulation ensures that a symmetric range of layers is selected around the midpoint of the model. The parameter \textit{mask\_cutoff} defines the width of the interval (approximately one-third of the total number of decoder layers, rounded to an even number), while $lb$ and $ub$ mark the lower and upper bounds of the chosen layer indices, meaning layers with the indices in $[lb,ub]$ are unmasked.

\subsubsection{Sequence repetition}
\label{app:seq_rep}
Repetition is implemented by concatenating a given input tensor $k$ times. This is done before the model's forward pass, meaning that padding tokens are also repeated. We also experimented with padding that leaves no padding tokens between each repetition but found no significant difference in performance, likely because rotary positional embeddings encode relative positional information \cite{su2024roformer}.

\subsection{Tokenization, optimization, and compute} 
\label{sec:experiment_details}

RoBERTa requires tokenization with added prefix space, and its tokenizer was the only one enabled with it. We adjust the cross-entropy loss to consider only the first token of each tokenized word from the input sequence. Maximum length for a tokenized sequence is set to $256$ tokens since no sequence among the tested datasets exceeds it. All of the reports are computed from five runs with $seeds = {5, 29, 42, 81, 123}$. Tokenized sequences are padded to the length of the longest sequence in a given batch.

In our QLoRA usage, we quantize all models with four-bit precision. Model parameters are stored using a normalized four-bit floating-point format, $nf4$. Computation is performed in a half-precision $fp16$ format. No double quantization is performed. This configuration reduces the memory footprint when training models, allowing us to efficiently perform training on the resources available to us. 

The total GPU usage for all experiments amounts to 2800 hours on \textit{Ampere A100} GPU. 

\begin{table}[ht]
\centering
\setlength{\tabcolsep}{3.5pt}
\begin{tabular}{ll}
\toprule
Model family & Model identifier          \\ \midrule
Gemma        & google/gemma-7b           \\ 
Gemma2      & google/gemma-2-2b         \\ 
Gemma2      & google/gemma-2-9b         \\ 
Mistral      & mistralai/Mistral-7B-v0.3 \\ 
Qwen3       & Qwen/Qwen3-0.6B           \\ 
Qwen3       & Qwen/Qwen3-1.7B           \\ 
Qwen3       & Qwen/Qwen3-4B             \\ 
Qwen3       & Qwen/Qwen3-8B             \\ 
RoBERTa      & FacebookAI/roberta-large       \\
ModernBERT   & answerdotai/ModernBERT-large   \\ \bottomrule
\vspace{-1em}
\end{tabular}
\caption{Models used and their Hugging Face identifiers}
\label{tab:models}
\end{table}

\begin{table}[ht]
  \centering
  \setlength{\tabcolsep}{3.5pt}
\resizebox{\linewidth}{!}{
  \begin{tabular}{@{}crrrrr@{}}
    \toprule
    \multicolumn{1}{c}{\multirow{1}{*}{\textbf{Dataset}}} & \multicolumn{1}{c}{\textbf{Train}} & \multicolumn{1}{r}{\textbf{Valid}} & \multicolumn{1}{r}{\textbf{Test}} & \multicolumn{1}{r}{\textbf{Total}} & \multicolumn{1}{r}{\textbf{\#Cl.}} \\
    \midrule
    CoNLL03 & \num{14041}  & \num{3250} & \num{3453} & \num{20744} & 4 \\
    ACE05 & \num{14672} & 873 & 711 & \num{16256} & 33 \\  
    Rest14 & \num{2737} & \num{304} & \num{800} & \num{3841} & 4 \\  
    NLU++ & \num{2152} & 309 & 619 & \num{3080} & 17 \\  
    \bottomrule
  \end{tabular}
  }
  \caption{Dataset statistics: the number of sentences per split, the total number of sentences, and the total number of class labels.}
  \label{tab:dataset_stats}
\end{table}
\begin{figure*}[ht]
    \centering
    \includegraphics[width=1.0\linewidth]{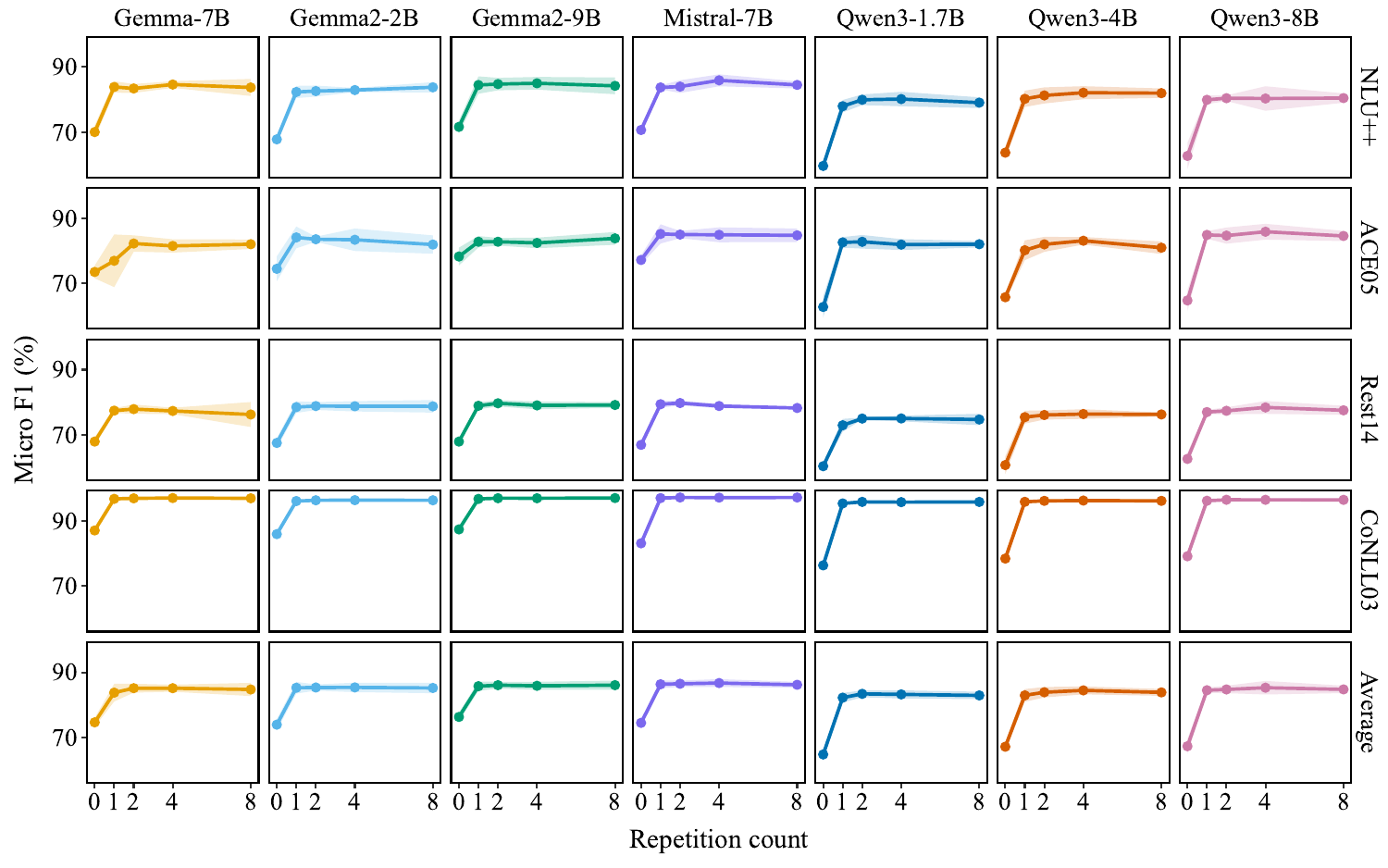}
    \caption{Micro F1 performance for SR across models and validation splits of each dataset. Subplots are aligned in a 5x7 grid, where rows and columns correspond to datasets and models, respectively. In the fifth row, we display the performance averaged over all the datasets. Shaded bands denote $\pm95\%$ CI. All results are averages of five runs.}
    \label{fig:repeat_k_valid}
\end{figure*}
\section{Full experimental results}
\label{app:full_results}
In \autoref{tab:full} we report the full results for all the considered models, datasets, and adaptation strategies. In \autoref{fig:repeat_k_valid} we report the results for decoder-only models with sequence repetition on validation sets of all datasets.

\begin{table*}[ht]
\centering
\small
\renewcommand{\arraystretch}{1.2}
\begin{tabular*}{\textwidth}{l@{\extracolsep{\fill}}lcccccc}
\toprule
\textbf{Model} & \textbf{Method} & \textbf{$r$} & \textbf{NLU++} & \textbf{ACE05} & \textbf{Rest14} & \textbf{CoNLL03} & \textbf{Average} \\
\midrule
\multirow{7}{*}{\textbf{Gemma-7B}} & \multirow{5}{*}{Sequence repetition} & 0 & $69.48_{{\pm 2.29}}$ & $69.40_{{\pm 2.64}}$ & $73.24_{{\pm 0.49}}$ & $81.87_{{\pm 0.33}}$ & $73.50_{{\pm 0.89}}$ \\
 & & 1 & $79.70_{{\pm 2.71}}$ & $71.72_{{\pm 3.45}}$ & $82.16_{{\pm 1.79}}$ & $93.20_{{\pm 0.32}}$ & $81.69_{{\pm 1.19}}$ \\
 & & 2 & $80.44_{{\pm 2.30}}$ & $74.67_{{\pm 2.22}}$ & $82.60_{{\pm 1.15}}$ & $93.38_{{\pm 0.24}}$ & $82.77_{{\pm 0.85}}$ \\
 & & 4 & $81.29_{{\pm 2.23}}$ & $74.27_{{\pm 1.41}}$ & $82.40_{{\pm 0.80}}$ & $93.38_{{\pm 0.18}}$ & $82.84_{{\pm 0.69}}$ \\
 & & 8 & $79.34_{{\pm 2.04}}$ & $71.62_{{\pm 6.00}}$ & $81.79_{{\pm 1.76}}$ & $93.25_{{\pm 0.16}}$ & $81.50_{{\pm 1.64}}$ \\
 & Middle unmasking & - & $79.90_{{\pm 3.36}}$ & $72.15_{{\pm 5.54}}$ & $83.64_{{\pm 0.68}}$ & $91.46_{{\pm 4.93}}$ & $81.79_{{\pm 2.04}}$ \\
 & Full unmasking & - & $74.34_{{\pm 2.77}}$ & $73.08_{{\pm 6.61}}$ & $76.63_{{\pm 4.97}}$ & $92.75_{{\pm 0.16}}$ & $79.20_{{\pm 2.18}}$ \\
\midrule
\multirow{7}{*}{\textbf{Gemma2-2B}} & \multirow{5}{*}{Sequence repetition} & 0 & $65.43_{{\pm 2.05}}$ & $64.33_{{\pm 3.77}}$ & $72.00_{{\pm 0.66}}$ & $80.13_{{\pm 0.46}}$ & $70.47_{{\pm 1.09}}$ \\
 & & 1 & $78.07_{{\pm 1.00}}$ & $74.11_{{\pm 1.73}}$ & $82.24_{{\pm 0.62}}$ & $92.60_{{\pm 0.15}}$ & $81.75_{{\pm 0.52}}$ \\
 & & 2 & $79.98_{{\pm 1.21}}$ & $72.86_{{\pm 3.64}}$ & $83.03_{{\pm 1.03}}$ & $93.05_{{\pm 0.29}}$ & $82.23_{{\pm 1.00}}$ \\
 & & 4 & $79.27_{{\pm 1.68}}$ & $72.79_{{\pm 1.20}}$ & $82.26_{{\pm 1.08}}$ & $93.09_{{\pm 0.14}}$ & $81.85_{{\pm 0.58}}$ \\
 & & 8 & $78.91_{{\pm 0.90}}$ & $72.87_{{\pm 4.34}}$ & $82.50_{{\pm 0.74}}$ & $92.87_{{\pm 0.15}}$ & $81.78_{{\pm 1.12}}$ \\
 & Middle unmasking & - & $78.21_{{\pm 2.03}}$ & $72.30_{{\pm 3.99}}$ & $82.21_{{\pm 0.40}}$ & $92.60_{{\pm 0.32}}$ & $81.33_{{\pm 1.13}}$ \\
 & Full unmasking & - & $76.42_{{\pm 2.16}}$ & $74.33_{{\pm 4.20}}$ & $81.26_{{\pm 0.64}}$ & $92.82_{{\pm 0.36}}$ & $81.21_{{\pm 1.19}}$ \\
\midrule
\multirow{7}{*}{\textbf{Gemma2-9B}} & \multirow{5}{*}{Sequence repetition} & 0 & $68.99_{{\pm 1.94}}$ & $67.24_{{\pm 2.86}}$ & $72.82_{{\pm 1.27}}$ & $82.20_{{\pm 0.38}}$ & $72.81_{{\pm 0.93}}$ \\
 & & 1 & $77.76_{{\pm 1.52}}$ & $76.69_{{\pm 2.63}}$ & $82.46_{{\pm 0.64}}$ & $93.44_{{\pm 0.23}}$ & $82.58_{{\pm 0.78}}$ \\
 & & 2 & $78.83_{{\pm 2.47}}$ & $74.49_{{\pm 1.71}}$ & $83.51_{{\pm 0.89}}$ & $93.58_{{\pm 0.24}}$ & $82.60_{{\pm 0.79}}$ \\
 & & 4 & $78.50_{{\pm 1.55}}$ & $72.59_{{\pm 3.18}}$ & $83.14_{{\pm 1.20}}$ & $93.59_{{\pm 0.16}}$ & $81.95_{{\pm 0.94}}$ \\
 & & 8 & $78.17_{{\pm 1.34}}$ & $73.83_{{\pm 4.28}}$ & $83.04_{{\pm 0.22}}$ & $93.51_{{\pm 0.18}}$ & $82.14_{{\pm 1.12}}$ \\
 & Middle unmasking & - & $77.63_{{\pm 1.23}}$ & $75.91_{{\pm 2.93}}$ & $83.34_{{\pm 0.97}}$ & $93.17_{{\pm 0.36}}$ & $82.51_{{\pm 0.84}}$ \\
 & Full unmasking & - & $76.83_{{\pm 2.09}}$ & $75.79_{{\pm 3.70}}$ & $82.55_{{\pm 1.28}}$ & $93.44_{{\pm 0.13}}$ & $82.15_{{\pm 1.11}}$ \\
\midrule

\multirow{7}{*}{\textbf{Mistral-7B}} & \multirow{5}{*}{Sequence repetition} & 0 & $69.60_{{\pm 1.52}}$ & $69.77_{{\pm 1.66}}$ & $72.51_{{\pm 1.06}}$ & $77.88_{{\pm 0.60}}$ & $72.44_{{\pm 0.64}}$ \\
 & & 1 & $79.93_{{\pm 1.46}}$ & $74.89_{{\pm 3.73}}$ & $83.84_{{\pm 1.03}}$ & $93.67_{{\pm 0.09}}$ & $83.08_{{\pm 1.03}}$ \\
 & & 2 & $80.71_{{\pm 0.72}}$ & $74.83_{{\pm 6.57}}$ & $83.54_{{\pm 0.38}}$ & $93.66_{{\pm 0.23}}$ & $83.19_{{\pm 1.66}}$ \\
 & & 4 & $80.28_{{\pm 1.78}}$ & $77.79_{{\pm 5.92}}$ & $83.31_{{\pm 1.57}}$ & $93.79_{{\pm 0.19}}$ & $83.79_{{\pm 1.59}}$ \\
 & & 8 & $81.05_{{\pm 3.25}}$ & $75.36_{{\pm 3.73}}$ & $84.18_{{\pm 0.37}}$ & $93.70_{{\pm 0.22}}$ & $83.57_{{\pm 1.24}}$ \\
 & Middle unmasking & - & $79.35_{{\pm 0.52}}$ & $77.56_{{\pm 3.31}}$ & $83.00_{{\pm 0.89}}$ & $93.48_{{\pm 0.12}}$ & $83.35_{{\pm 0.87}}$ \\
 & Full unmasking & - & $80.23_{{\pm 1.24}}$ & $76.26_{{\pm 3.77}}$ & $83.18_{{\pm 0.84}}$ & $93.69_{{\pm 0.24}}$ & $83.34_{{\pm 1.02}}$ \\
\midrule
\multirow{7}{*}{\textbf{Qwen3-1.7B}} & \multirow{5}{*}{Sequence repetition} & 0 & $59.58_{{\pm 2.58}}$ & $55.25_{{\pm 3.08}}$ & $65.71_{{\pm 1.18}}$ & $70.98_{{\pm 0.87}}$ & $62.88_{{\pm 1.07}}$ \\
 & & 1 & $74.53_{{\pm 1.91}}$ & $69.62_{{\pm 2.56}}$ & $77.76_{{\pm 0.50}}$ & $90.85_{{\pm 0.44}}$ & $78.19_{{\pm 0.82}}$ \\
 & & 2 & $76.67_{{\pm 1.53}}$ & $70.66_{{\pm 3.57}}$ & $79.25_{{\pm 0.26}}$ & $91.53_{{\pm 0.26}}$ & $79.53_{{\pm 0.98}}$ \\
 & & 4 & $77.75_{{\pm 1.43}}$ & $70.18_{{\pm 3.42}}$ & $79.26_{{\pm 1.03}}$ & $91.56_{{\pm 0.41}}$ & $79.69_{{\pm 0.97}}$ \\
 & & 8 & $76.08_{{\pm 1.57}}$ & $70.75_{{\pm 4.02}}$ & $78.95_{{\pm 1.29}}$ & $91.78_{{\pm 0.29}}$ & $79.39_{{\pm 1.13}}$ \\
 & Middle unmasking & - & $73.67_{{\pm 1.46}}$ & $66.71_{{\pm 3.87}}$ & $75.71_{{\pm 1.02}}$ & $90.31_{{\pm 0.42}}$ & $76.60_{{\pm 1.07}}$ \\
 & Full unmasking & - & $67.28_{{\pm 4.24}}$ & $68.68_{{\pm 3.39}}$ & $75.32_{{\pm 1.99}}$ & $90.27_{{\pm 0.45}}$ & $75.39_{{\pm 1.45}}$ \\
\midrule
\multirow{7}{*}{\textbf{Qwen3-4B}} & \multirow{5}{*}{Sequence repetition} & 0 & $64.98_{{\pm 2.31}}$ & $59.00_{{\pm 4.14}}$ & $68.26_{{\pm 0.50}}$ & $72.83_{{\pm 0.43}}$ & $66.27_{{\pm 1.20}}$ \\
 & & 1 & $77.64_{{\pm 1.13}}$ & $73.53_{{\pm 3.92}}$ & $80.14_{{\pm 1.88}}$ & $91.90_{{\pm 0.51}}$ & $80.80_{{\pm 1.13}}$ \\
 & & 2 & $78.43_{{\pm 1.43}}$ & $70.72_{{\pm 5.46}}$ & $80.85_{{\pm 0.87}}$ & $92.19_{{\pm 0.18}}$ & $80.55_{{\pm 1.43}}$ \\
 & & 4 & $78.12_{{\pm 2.33}}$ & $69.69_{{\pm 4.17}}$ & $81.16_{{\pm 0.79}}$ & $92.30_{{\pm 0.17}}$ & $80.32_{{\pm 1.21}}$ \\
 & & 8 & $77.97_{{\pm 1.29}}$ & $72.58_{{\pm 5.20}}$ & $81.22_{{\pm 0.48}}$ & $92.33_{{\pm 0.28}}$ & $81.02_{{\pm 1.35}}$ \\
 & Middle unmasking & - & $74.56_{{\pm 0.81}}$ & $69.73_{{\pm 2.91}}$ & $79.60_{{\pm 0.41}}$ & $91.87_{{\pm 0.34}}$ & $78.94_{{\pm 0.77}}$ \\
 & Full unmasking & - & $73.79_{{\pm 1.17}}$ & $70.59_{{\pm 4.48}}$ & $77.76_{{\pm 0.62}}$ & $91.17_{{\pm 0.17}}$ & $78.33_{{\pm 1.17}}$ \\
\midrule
\multirow{7}{*}{\textbf{Qwen3-8B}} & \multirow{5}{*}{Sequence repetition} & 0 & $64.20_{{\pm 3.17}}$ & $57.27_{{\pm 5.61}}$ & $68.83_{{\pm 1.71}}$ & $73.27_{{\pm 0.36}}$ & $65.89_{{\pm 1.67}}$ \\
 & & 1 & $76.38_{{\pm 2.82}}$ & $69.83_{{\pm 8.10}}$ & $81.47_{{\pm 0.53}}$ & $92.24_{{\pm 0.34}}$ & $79.98_{{\pm 2.15}}$ \\
 & & 2 & $77.88_{{\pm 1.56}}$ & $71.41_{{\pm 4.69}}$ & $82.26_{{\pm 0.20}}$ & $92.81_{{\pm 0.31}}$ & $81.09_{{\pm 1.24}}$ \\
 & & 4 & $77.61_{{\pm 2.52}}$ & $72.11_{{\pm 5.84}}$ & $82.27_{{\pm 1.11}}$ & $92.74_{{\pm 0.26}}$ & $81.18_{{\pm 1.62}}$ \\
 & & 8 & $77.53_{{\pm 1.21}}$ & $71.96_{{\pm 6.01}}$ & $81.96_{{\pm 0.80}}$ & $93.00_{{\pm 0.10}}$ & $81.11_{{\pm 1.55}}$ \\
 & Middle unmasking & - & $75.06_{{\pm 1.18}}$ & $71.45_{{\pm 2.97}}$ & $80.17_{{\pm 0.74}}$ & $92.23_{{\pm 0.38}}$ & $79.73_{{\pm 0.83}}$ \\
 & Full unmasking & - & $74.45_{{\pm 2.72}}$ & $69.01_{{\pm 3.11}}$ & $80.01_{{\pm 0.42}}$ & $91.99_{{\pm 0.29}}$ & $78.86_{{\pm 1.04}}$ \\
\midrule
{\textbf{ModernBERT}} & - & - & $70.50_{{\pm 3.24}}$ & $68.30_{{\pm 4.18}}$ & $75.51_{{\pm 1.01}}$ & $90.02_{{\pm 0.23}}$ & $76.08_{{\pm 1.35}}$ \\
{\textbf{RoBERTa}} & - & - & $72.22_{{\pm 1.68}}$ & $68.96_{{\pm 8.04}}$ & $80.30_{{\pm 0.78}}$ & $92.64_{{\pm 0.12}}$ & $78.53_{{\pm 2.06}}$ \\
\bottomrule
\end{tabular*}
\caption{Micro F1 scores for all models, methods, and repetition counts. Standard deviations are subscripted.}
\label{tab:full}
\end{table*}

\section{AI use statement}
During the preparation of this manuscript, we used AI assistance for language editing, rephrasing, and spell-checking.

\end{document}